\documentclass{article}

\usepackage[accepted]{icml2024}

\usepackage{microtype}
\usepackage{graphicx}
\usepackage{subfigure}
\usepackage{booktabs}
\usepackage{hyperref}
\usepackage{amsmath}
\usepackage{amssymb}
\usepackage{mathtools}
\usepackage{amsthm}
\usepackage[capitalize,noabbrev]{cleveref}
\usepackage{hyperref}
\usepackage{url}
\usepackage{booktabs}
\usepackage{graphicx}
\usepackage{subcaption}
\usepackage{cleveref}
\usepackage{amsmath,amssymb}
\usepackage[export]{adjustbox}[2011/08/13]

\theoremstyle{plain}

\theoremstyle{definition}

\theoremstyle{remark}

\begin{document}

\twocolumn[

\icmltitle{Mamba-PTQ: Outlier Channels in Recurrent Large Language Models}

\icmltitlerunning{Mamba-PTQ: Outlier Channels in Recurrent Large Language Models}

\begin{icmlauthorlist}
\icmlauthor{Alessandro Pierro*}{intel,lmu}
\icmlauthor{Steven Abreu*}{intel,groningen}
\end{icmlauthorlist}

\icmlaffiliation{intel}{Neuromorphic Computing Lab, Intel Labs, Neubiberg, Germany}
\icmlaffiliation{lmu}{Institute of Informatics, LMU Munich, Germany}
\icmlaffiliation{groningen}{CogniGron Center \& Bernoulli Institute, University of Groningen, Groningen, Netherlands}

\icmlcorrespondingauthor{Alessandro Pierro}{alessandro.pierro@intel.com}

\icmlkeywords{Recurrent Neural Networks, State Space Models, Language Modeling, Quantization}

\vskip 0.3in
]

\printAffiliationsAndNotice{\icmlEqualContribution}

\begin{abstract}
Modern recurrent layers are emerging as a promising path toward edge deployment of foundation models, especially in the context of large language models (LLMs).
Compressing the whole input sequence in a finite-dimensional representation enables recurrent layers to model long-range dependencies while maintaining a constant inference cost for each token and a fixed memory requirement.
However, the practical deployment of LLMs in resource-limited environments often requires further model compression, such as quantization and pruning.
While these techniques are well-established for attention-based models, their effects on recurrent layers remain underexplored.

In this preliminary work, we focus on post-training quantization for recurrent LLMs and show that Mamba models exhibit the same pattern of outlier channels observed in attention-based LLMs.
We show that the reason for difficulty of quantizing SSMs is caused by activation outliers, similar to those observed in transformer-based LLMs.
We report baseline results for post-training quantization of Mamba that do not take into account the activation outliers and suggest first steps for outlier-aware quantization.
\end{abstract}

\section{Introduction}
\label{sec:introduction}

Attention-based models, also known as Transformers \cite{vaswani_attention_2023}, constitute the current state-of-the-art backbone for large language models (LLMs)~\citep{brown_language_2020}.
However, their powerful modeling capabilities come with significant computational requirements, resulting in high inference costs and limiting the deployment on edge and low-power devices.
Novel recurrent neural network (RNN) architectures, informed mainly by recent work on state space models (SSMs) \cite{gu_hippo_2020,gu_efficiently_2022}, are now emerging as promising alternatives for sequence modeling tasks, either in isolation \cite{poli_hyena_2023,gu_mamba_2023,peng_rwkv_2023} or as hybrid models interleaving recurrent and attention blocks \cite{de_griffin_2024,lieber_jamba_2024,botev_recurrentgemma_2024}.
In particular, RNNs compress the input sequence into a finite-dimensional representation, decoupling the computational and memory cost of each token's forward pass from the sequence's length.
Hence, they provide better scalability to long context scenarios than vanilla self-attention, which scales quadratically with sequence length.

However, similarly to Transformers, deploying recurrence-based LLMs at scale or in resource-constrained environments requires advanced model optimization techniques, such as quantization, pruning, and knowledge distillation. 
While applying these techniques starts to be well understood in the context of attention-based LLMs, model optimization for recurrent and hybrid architectures remains an important yet underexplored topic.
In this paper, we focus on quantization and analyze its impact on Mamba \cite{gu_mamba_2023} model family, drawing connections to previous work on quantized LLMs.

\section{Quantization and outlier channels in LLMs}
\label{sec:background}

\begin{figure*}
    \centering
    \includegraphics[width=0.7\textwidth,center]{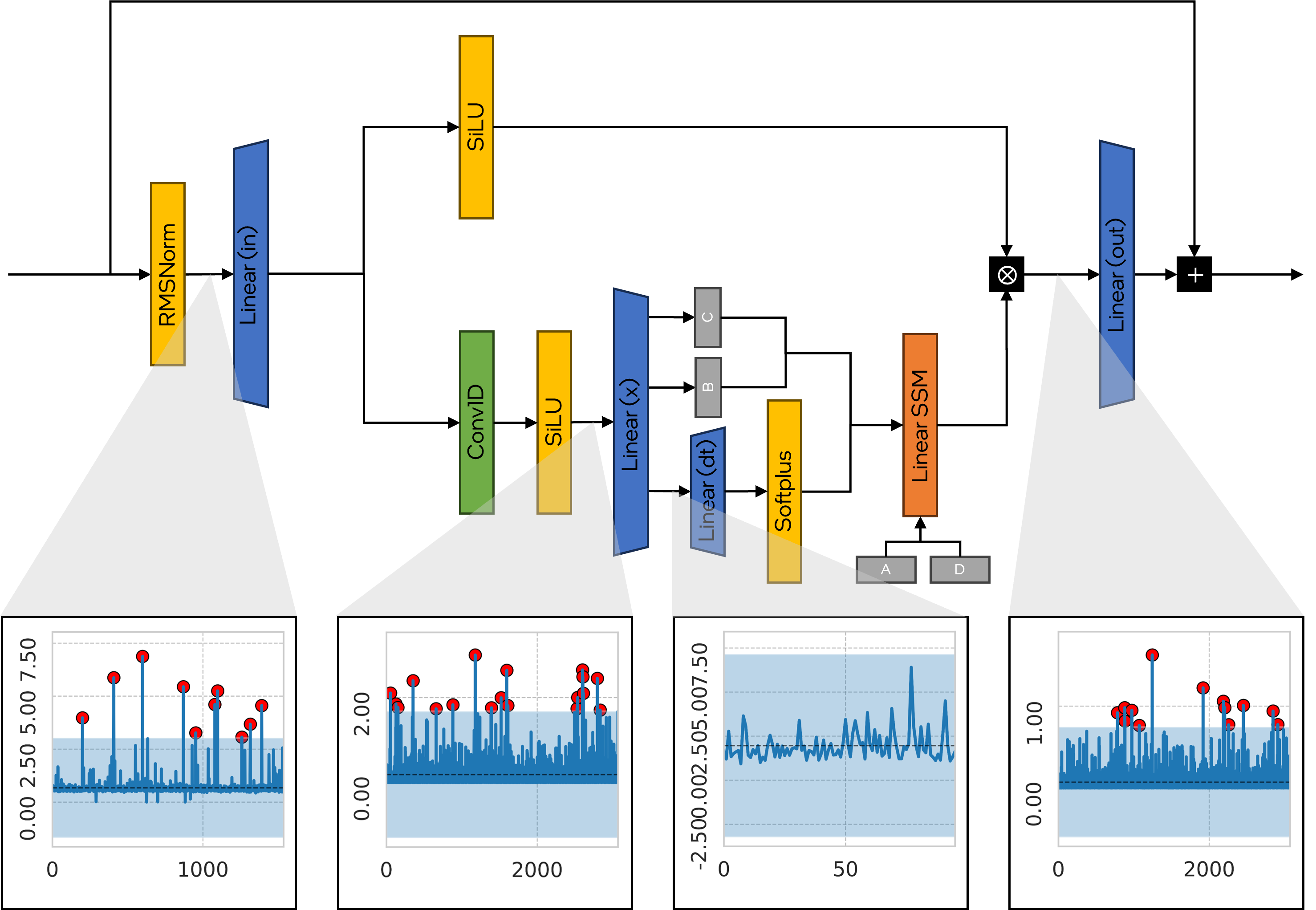}
    \caption{Architecture diagram of the Mamba block and details on the absolute maximum activation (on the y-axis) across channels (x-axis), measured on a subset of WikiText-2 \cite{merity_pointer_2016} for Mamba-130m. Shaded regions account for six standard deviations.}
    \label{fig:overview}
\end{figure*}

Quantization is a compression technique that reduces the numerical precision of a model's weights and activations to integer datatypes in order to facilitate inference \cite{jacob_quantization_2017}.
We adopt symmetric per-tensor quantization: given a tensor $\mathbf{x}$ and a bit precision $n$, its quantized representation is computed as:
\begin{equation}
    \bar{\mathbf{x}}_{n} = \left\lfloor \frac{(2^{n-1}-1) \mathbf{x}}{\max | \mathbf{x} |} \right\rceil = \left\lfloor s_x \mathbf{x}\right\rceil
\end{equation}
where the \textit{quantization scale} $s_x$ is a scalar.
The benefits of quantization for inference efficiency are twofold.
Firstly, weight quantization reduces the memory footprint of the model,  which is especially beneficial in the memory-bound regime of autoregressive generation.
Secondly, when both weights and activations are quantized, matrix multiplications can be offloaded to the integer processing units, which typically offer higher throughput and energy efficiency than floating point units.

Most state-of-the-art techniques for quantizing LLMs are based on the empirical observation of \textit{outlier channels} \cite{bondarenko_understanding_2021}, a small percentage of model dimensions with a dynamic range that is consistently larger than the rest.
This phenomenon complicates activation quantization since the large abs max values from the outlier channels deteriorate the effective bit precision of the remaining channels.
A possible solution would be maintaining a different quantization scale for each channel, which is not hardware-friendly on current GPU architectures \cite{xiao_smoothquant_2024}.
Various strategies have been proposed to circumvent this issue. For instance, some methods treat outlier channels separately, either by maintaining them in floating point format \cite{dettmers_llmint8_2022} or by representing them with two integer channels each \cite{zhang_flattenquant_2024}. Other approaches modify the transformer architecture to prevent the emergence of outliers \cite{bondarenko_quantizable_2023}, while some partially shift the quantization difficulty to the weights, thereby mitigating the impact of outliers \cite{xiao_smoothquant_2024}.

We make the first steps towards post-training quantization for recurrent LLMs, focusing on the Mamba \cite{gu_mamba_2023} model family.
We analyse the activation patterns of Mamba to assess the presence of outliers, which we define as those channels having an absolute maximum activation beyond six standard deviations from the layer mean, following prior practice \cite{bondarenko_understanding_2021}.
\autoref{fig:overview} reports the pre-activations of the linear block of a layer from Mamba-130m (similar results were observed for the other model sizes), measured running the model on a subset of WikiText-2 \cite{merity_pointer_2016}.
We observe distinct outlier patterns.
The pre-activations of the three largest linear layers (\textit{in}, \textit{x}, and \textit{out}), consistently with what was observed for attention-based LLMs, show outliers accounting for less than $1\%$ of channels.
However, while the outliers of the first linear block are mostly consistent across layers, the remaining two blocks exhibit no regular behavior.
The linear layer projecting the SSM's time steps (\textit{dt}) shows almost no outliers.
Similarly to \citep{dettmers_llmint8_2022}, we further assess the importance of the outlier channels for the model's predictions by evaluating the impact of zeroing out the outliers on downstream accuracy.
For Mamba-130m and Mamba-2.8b, we observe a drop in average accuracy of $12.61\%$ and $17.49\%$, respectively, suggesting that these channels play a significant role in the model dynamics.
Extended results are available in the Appendix in \autoref{tab:outlier-removal}.

\section{Method}
\label{sec:methodology}

\subsection{Mamba model}

Existing state space models are described by the following dynamics:
\begin{align}
    x^k &= A x^{k-1} + B u^k \\
    y^k &= C x^k + D u^k
\end{align}
where $A$ is the recurrent matrix, $B$ is the input matrix, $C$ is the output matrix, and $D$ is the residual matrix from the input to the output. State space models of this form must treat every input token equally, as the input matrix $B$ is fixed, such that the SSM cannot focus on or ignore specific tokens. This is a major shortcoming compared to transformer architectures, whose attention mechanism allows for such interactions and thus limits the performance of SSMs, especially on language tasks. 

The key innovation of Mamba over previous SSMs is its ability to perform such content-based reasoning. By letting Mamba's parameters depend on the input, the model effectively gains the ability to filter out irrelevant information so that the relevant context can be compressed more efficiently into the hidden state. Specifically, the parameters for Mamba's SSM block are obtained by:
\begin{align}
    B_t, \Delta_t, C_t &= W^{proj} u_t \label{eq:ssm-matrices}\\
    \overline{\Delta_t} &= \sigma^+(W^{dt} \Delta_t) \\
    \overline{A_t} &= \exp \left( -\exp \left( A^{log} \overline{\Delta_t} \right) \right) \\
    \overline{B_t} &= \overline{\Delta_t} \odot B_t
\end{align}
where $u_t$ is the input to the SSM block, $W^{proj}$, $A^{log}$ and $W^{dt}$ are time-invariant weight matrices, $\sigma^+$ denotes the softplus function and $\odot$ denotes element-wise multiplication. The weight matrices $B_t$ and $C_t$ are thus directly dependent on the input $u_t$, whereas the recurrent matrix $A_t$ is dependent on the input $u_t$ only through the input-dependent timescale parameter $\Delta_t$. The hidden state $h_t$ and output $y_t$ of the SSM block is then computed as:
\begin{align}
    h_t &= \overline{A_t} \odot h_{t-1} + \overline{B_t} \odot u_t \\
    y_t &= C_t h_t + D_t \odot u_t
\end{align}
As shown in Figure \ref{fig:overview}, each layer in the Mamba architecture also includes additional gating, nonlinearities, normalization, causal convolution, and linear blocks.

\subsection{Baseline quantization}

In order to quantize Mamba, we distinguish between Mamba's pre-trained weights and its activations. Importantly, due to the input-dependent parameterization, we consider only input-independent parameters as weights, such as $A^{log}$, while we consider input-dependent parameters like $\overline{A_t}$ as activations. 

We adopt symmetric, per-tensor quantization for weights and activations as described in \autoref{sec:background}, using the absolute maximum (absmax) of the tensor for calibration.

For our experiments using naive quantization on the activations, we quantize the output from all linear layers (including the matrices $B_t$, $\Delta_t$, $C_t$ from \autoref{eq:ssm-matrices}), but we do not quantize the effective weight matrices $\overline{A_t}, \overline{\Delta_t}, \overline{B_t}$. We further do not quantize the output from the SSM block $y_t$ but only quantize the output from the downstream out projection linear layer.

We use standard notation to denote quantization with $n$-bit integers for weights as W$n$ and quantization with $n$-bit integers for activations as A$n$. For example, 8-bit weight quantization and 4-bit activation quantization is denoted by W8A8. 

\subsection{Outlier-aware quantization (e.g., SmoothQuant)}

The naive absmax quantization is sensitive to outliers. A large value in the tensor $\mathbf{x}$ will yield a small scale $s_x = \frac{2^{n-1}-1}{\max |\mathbf{x}|}$, thus leading to larger rounding errors for the same $n$-bit quantization precision. As discussed in \autoref{sec:background}, outliers (particularly in activations) are the subject of research in LLM quantization. 

Most notably, the SmoothQuant method proposed by Xiao \emph{et al.} \cite{xiao_smoothquant_2024} exploits the fact that outliers exist in activations but not in the weights. SmoothQuant smooths the activation outliers by partially taking them into the preceding weights. Because activation outliers typically persist in the same activation channels, a weight matrix with per-channel quantization can absorb part of the quantization difficulty from the subsequent activations. As such, SmoothQuant introduces a per-channel smoothing factor $s \in \mathbb{R}^{C_i}$ where $C_i$ is the dimension of the activations $X$ and, equivalently, the number of output channels of the weight matrix $W$. This smpoothing factor is used to scale the weights and activations:
\begin{align}
    (\mathbf{X} \text{diag}(s)^{-1}) \cdot (\text{diag}(s) \textbf{W}) = \hat{\mathbf{X}} \hat{\mathbf{W}}
\end{align}
The aim is to choose a smoothing factor $s$ so that $\hat{\mathbf{X}} = \mathbf{X} \text{diag}(s)^{-1}$ is easy to quantize. However, simply choosing $s_j=\max (|\mathbf{X}_j|)$ where $j=1,\ldots,C_i$ to minimize the difficulty in quantization activations, will push all these difficulties into the weights. On the other hand, we can choose $s_j = 1 / \max (|\mathbf{W}_j|)$ to move all the quantization difficulty from the weights into the activations. The authors propose a new hyperparameter, the migration strength $\alpha$, to control how much difficulty we want to migrate from activations to weights, using the equation:
\begin{align}
    s_j = \frac{\max (| \mathbf{X}_j |)^\alpha}{\max (|\mathbf{W}_j|)^{1-\alpha}}
\end{align}
where a smaller $\alpha$ will leave more difficulty with the activations, and a larger $\alpha$ will migrate more difficulty to the weights. The authors suggest to use a default value of $\alpha=0.5$ and a larger $\alpha$ for models where activation outliers are more significant such that more quantization difficulty is moved into the weights.

\begin{table*}
\centering
\caption{One-shot accuracy on downstream tasks for Mamba-1.4b across different quantization configurations.}
\begin{tabular}{lccccccc}
\toprule
 & \textbf{LAMBADA} & \textbf{HellaSwag} & \textbf{PIQA} & \textbf{WinoGrande} & \textbf{RTE} & \textbf{COPA} \\
\midrule
\quad Baseline      & $64.95\%$ & $59.11\%$ & $74.16\%$ & $61.4\%$  & $48.01\%$ & $79\%$ \\
\midrule
\quad W8 (mlp) &$64.43\%$ & $44.91\%$ & $74.32\%$ & $60.06\%$ & $48.01\%$ & $77.00\%$ \\
\quad W8 (all)      & $63.01\%$ & $44.71\%$ & $73.07\%$ & $60.06\%$ & $51.62\%$ & $76.00\%$ \\
\quad W4 (mlp) & $0.02\%$ & $25.70\%$ & $52.29\%$ & $51.54\%$ & $52.35\%$ & $56.00\%$ \\
\quad W4 (all) & $0.00\%$ & $25.72\%$ & $52.39\%$ & $50.99\%$ & $55.23\%$ & $64.00\%$ \\
\quad W8A8 (mlp) & $63.11\%$ & $44.42\%$ & $73.01\%$ & $60.06\%$ & $51.62\%$ & $76\%$ \\
\quad W8A8 (all) & $55.35\%$ & $43.84\%$ & $70.24\%$ & $54.3\%$ & $52.71\%$ & $75\%$ \\
\bottomrule
\end{tabular}
\label{tab:lm_results}
\end{table*}

\section{Experiments}
\label{sec:experiments}

\begin{figure}
    \centering
    \includegraphics[width=\linewidth]{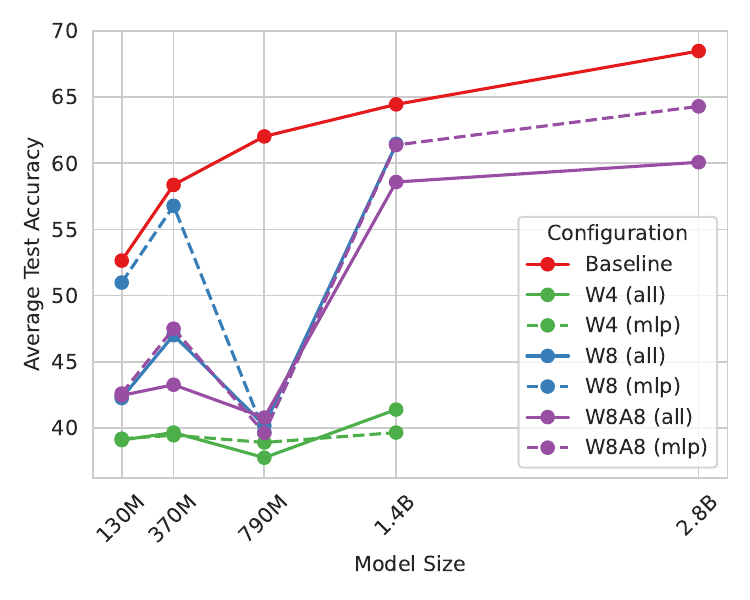}
    \caption{Average one-shot accuracy on downtream tasks across model sizes for Mamba with different quantization configurations. The accuracy is averaged over all tasks shown in \autoref{tab:lm_results}.}
    \label{fig:results}
\end{figure}

\subsection{Experimental setup}

We assess the impact of different quantization configurations on the zero-shot accuracy of six downstream tasks: LAMBADA \cite{paperno_lambada_2016}, HellaSwag \cite{zellers_hellaswag_2019}, PIQA \cite{bisk_piqa_2019}, WinoGrande \cite{sakaguchi_winogrande_2019}, RTE \cite{wang_glue_2019}, and COPA \cite{roemmele_choice_nodate}.
We explicitly neglect perplexity benchmarking, since prior work noted how it may not be informative of the actual task performance degradation \cite{sun2021longrange}.

We run three different experimental conditions:
\begin{enumerate}
    \item To assess the importance of outlier channels, we analyze the impact of removing outlier channels on downstream task accuracy. 
    \item We then analyze the effect of naive quantization of only the pre-trained weights on downstream task accuracy. 
    \item Finally, we analyze the effect of quantization on the pre-trained weights, as well as the activations, without accounting for activation outliers.
\end{enumerate}
Full results for the effect of Experiment \textbf{1} for removing outlier channels are found in \autoref{tab:outlier-removal} in the Appendix. We present an overview of the findings from Experiment \textbf{2} and \textbf{3} in \autoref{tab:lm_results} on the Mamba-1.4b model, while results for all other model sizes are presented in \autoref{tab:lm_results_appendix} in the Appendix.

\section{Discussion}
\label{sec:discussion}

In this preliminary work, we make the first steps towards post-training quantization of Mamba, in order to inform future edge deployments of recurrent LLMs based on selective state space models such as Mamba.
We have shown that the difficulty of quantizing Mamba is caused by activation outliers, similar to those observed in transformer-based LLMs. We presented baseline results for post-training quantization of Mamba that does not take into account the activation outliers and a first proposal for outlier-aware quantization of Mamba.

\subsection{Future work}

As this area is under rapid development, several opportunities exist to extend this work.
Firstly, a similar analysis could be performed on other recurrent LLMs, such as the RWKV family \cite{peng_rwkv_2023}, the novel Mamba-2 architecture \cite{dao2024transformers}, or hybrid models such as Griffin \cite{de_griffin_2024} and RecurrentGemma \cite{botev_recurrentgemma_2024}.
Secondly, additional work should be done to convert the SSM dynamics fully to integer operations, as previously demonstrated by \citep{blouw2021hardware}, and explore the use of quantized activations.
Lastly, it will be interesting to see how quantized recurrent LLMs perform at the edge in energy-constrained scenarios for real-time multimodal processing \cite{shrestha_efficient_2024}, as the specific of the hardware architecture could provide additional guidance on model compression requirements.

\section*{Acknowledgements}

We thank Jonathan Timcheck, Philipp Stratmann, and Sumit Shrestha for helpful comments and discussions.

\bibliography{example_paper}
\bibliographystyle{icml2024}

\newpage
\appendix
\onecolumn

\section{Additional results}

Herein we present all additional experimental results from the experiments presented in this paper.

\subsection{Impact of removing outlier channels}

Table \ref{tab:outlier-removal} shows the accuracy on all evaluated tasks for the Mamba-130m model and Mamba-2.8B model, with different rows indicating the outlier removal specific to particular layers, or across the entire model.

\begin{table*}[h]
\centering
\caption{Impact of removing outlier channels on downstream task accuracy.}
\begin{tabular}{lccccccc}
\toprule
\textbf{Model} & \textbf{LAMBADA} & \textbf{HellaSwag} & \textbf{PIQA} & \textbf{WinoGrande} & \textbf{RTE} & \textbf{COPA} & Avg. \\
\midrule
\textbf{Mamba-130m} \\
\quad Baseline          & $44.25\%$ & $35.25\%$ & $64.47\%$ & $52\%$    & $54.87\%$ & $65\%$ & $52.64\%$ \\
\quad Linear(in) only   & $0\%$     & $25.61\%$ & $53.75\%$ & $51.14\%$ & $52.71\%$ & $53\%$ & $39.37\%$ \\
\quad Linear(x) only    & $26.88\%$ & $26.73\%$ & $58.49\%$ & $49.57\%$ & $55.6\%$  & $63\%$ & $46.71\%$ \\
\quad Linear(dt) only   & $44.25\%$ & $30.80\%$ & $64.47\%$ & $52.09\%$ & $54.87\%$ & $65\%$ & $51.91\%$ \\
\quad Linear(out) only  & $32.45\%$ & $30.42\%$ & $65.34\%$ & $53.35\%$ & $53.43\%$ & $70\%$ & $50.83\%$ \\
\quad All               & $0\%$     & $25.83\%$ & $54.52\%$ & $53.12\%$ & $52.71\%$ & $54\%$ & $40.03\%$ \\
\midrule
\textbf{Mamba-2.8B} \\
\quad Baseline          & $69.24\%$ & $66.16\%$ & $75,24\%$ & $63.46\%$ & $52.71\%$ & $84\%$    & $68.46\%$ \\
\quad All               & $7.45\%$  & $49.43\%$      & $66.92\%$  & $58.25\%$  & $53.79\%$  & $70\%$    & $50.97\%$ \\
\bottomrule
\end{tabular}
\label{tab:outlier-removal}
\end{table*}

\newpage
\subsection{Impact of quantization on downstream task accuracy}

Table \ref{tab:lm_results_appendix} shows the accuracy on all evaluated tasks for all Mamba models and all quantization configurations. 

\begin{table}[h]
\centering
\caption{One-shot accuracy on downstream tasks for the Mamba model family across different quantization configurations.}
\begin{tabular}{lcccccc}
\toprule
\textbf{Model} & \textbf{LAMBADA} & \textbf{HellaSwag} & \textbf{PIQA} & \textbf{WinoGrande} & \textbf{RTE} & \textbf{COPA} \\
\midrule
\textbf{Mamba-130m} \\
\quad Baseline & $44.25\%$ & $35.25\%$ & $64.47\%$ & $52\%$ & $54.87\%$ & $65\%$ \\
\quad W8 (mlp) & $42.48\%$ & $30.71\%$ & $64.09\%$ & $52.88\%$ & $52.71\%$ & $63.00\%$ \\
\quad W8 (all) & $5.53\%$ & $28.03\%$ & $58.11\%$ & $50.59\%$ & $47.29\%$ & $64.00\%$ \\
\quad W4 (mlp) & $0.00\%$ & $25.80\%$ & $51.74\%$ & $50.75\%$ & $49.82\%$ & $57.00\%$ \\
\quad W4 (all) & $0.00\%$ & $25.34\%$ & $53.43\%$ & $50.36\%$ & $52.35\%$ & $53.00\%$ \\
\quad W8A8 (mlp) & $5.72\%$ & $28.09\%$ & $57.83\%$ & $51.3\%$ & $47.65\%$ & $65\%$ \\
\quad W8A8 (all) & $4.31\%$ & $27.72\%$ & $56.47\%$ & $51.07\%$ & $50.18\%$ & $65\%$ \\
\midrule
\textbf{Mamba-370m} \\
\quad Baseline & $55.62\%$ & $46.48\%$ & $69.48\%$ & $55.49\%$ & $53.07\%$ & $70\%$ \\
\quad W8 (mlp) & $54.86\%$ & $37.17\%$ & $69.04\%$ & $55.80\%$ & $53.79\%$ & $70.00\%$ \\
\quad W8 (all) & $16.61\%$ & $31.58\%$ & $61.37\%$ & $51.38\%$ & $49.10\%$ & $72.00\%$ \\
\quad W4 (mlp) & $0.00\%$ & $25.23\%$ & $53.81\%$ & $50.28\%$ & $52.35\%$ & $55.00\%$ \\
\quad W4 (all) & $0.00\%$ & $25.72\%$ & $53.10\%$ & $51.38\%$ & $52.71\%$ & $55.00\%$ \\
\quad W8A8 (mlp) & $16.61\%$ & $36.87\%$ & $61.53\%$ & $51.22\%$ & $48.74\%$ & $70\%$ \\
\quad W8A8 (all) & $10.29\%$ & $31.04\%$ & $58.76\%$ & $50.99\%$ & $45.49\%$ & $63\%$ \\
\midrule
\textbf{Mamba-790m} \\
\quad Baseline & $61.71\%$ & $55.07\%$ & $72.14\%$ & $55.96\%$ & $55.23\%$ & $72\%$ \\
\quad W8 (mlp) & $2.64\%$ & $25.30\%$ & $54.13\%$ & $50.83\%$ & $53.07\%$ & $52.00\%$ \\
\quad W8 (all) & $1.20\%$ & $25.84\%$ & $54.08\%$ & $51.07\%$ & $55.96\%$ & $53.00\%$ \\
\quad W4 (mlp) & $0.00\%$ & $25.93\%$ & $53.05\%$ & $48.22\%$ & $46.21\%$ & $60.00\%$ \\
\quad W4 (all) & $0.00\%$ & $25.29\%$ & $50.98\%$ & $47.67\%$ & $46.57\%$ & $56.00\%$ \\
\quad W8A8 (mlp) & $1.44\%$ & $25.37\%$ & $54.30\%$ & $50.2\%$ & $53.43\%$ & $53\%$ \\
\quad W8A8 (all) & $0.95\%$ & $25.77\%$ & $54.73\%$ & $50.99\%$ & $55.23\%$ & $57\%$ \\
\midrule
\textbf{Mamba-1.4B} \\
\quad Baseline      & $64.95\%$ & $59.11\%$ & $74.16\%$ & $61.4\%$  & $48.01\%$ & $79\%$ \\
\quad W8 (mlp) &$64.43\%$ & $44.91\%$ & $74.32\%$ & $60.06\%$ & $48.01\%$ & $77.00\%$ \\
\quad W8 (all)      & $63.01\%$ & $44.71\%$ & $73.07\%$ & $60.06\%$ & $51.62\%$ & $76.00\%$ \\
\quad W4 (mlp) & $0.02\%$ & $25.70\%$ & $52.29\%$ & $51.54\%$ & $52.35\%$ & $56.00\%$ \\
\quad W4 (all) & $0.00\%$ & $25.72\%$ & $52.39\%$ & $50.99\%$ & $55.23\%$ & $64.00\%$ \\
\quad W8A8 (mlp) & $63.11\%$ & $44.42\%$ & $73.01\%$ & $60.06\%$ & $51.62\%$ & $76\%$ \\
\quad W8A8 (all) & $55.35\%$ & $43.84\%$ & $70.24\%$ & $54.3\%$ & $52.71\%$ & $75\%$ \\
\midrule
\textbf{Mamba-2.8B} \\
\quad Baseline & $69.24\%$ & $66.16\%$ & $75.24\%$ & $63.46\%$ & $52.71\%$ & $84\%$ \\
\quad W8A8 (mlp) & $64.64\%$ & $48.74\%$ & $73.72\%$ & $64.33\%$ & $56.32\%$ & $78\%$ \\
\quad W8A8 (all) & $51.39\%$ & $47.64\%$ & $70.24\%$ & $57.62\%$ & $54.51\%$ & $79\%$ \\
\bottomrule
\end{tabular}
\label{tab:lm_results_appendix}
\end{table}

\end{document}